\documentclass{vgtc}                          

\onlineid{1002}

\vgtccategory{Research}

\vgtcpapertype{Data Transformations}

\title{Ensemble Visualization With Variational Autoencoder}

\author{Cenyang Wu\thanks{e-mail: wuceny@stu.pku.edu.cn}\\ %
     \parbox{2.2in}{   \scriptsize \centering Institute of Medical Technology, Peking University Health Science Center\\National Institute of Health Data Science, Peking University} %
\and Qinhan Yu\thanks{e-mail: yuqinhan@stu.pku.edu.cn}\\ %
     \parbox{1.4in}{   \scriptsize \centering  Center for Machine Learning Research, Peking University}
\and Liang Zhou\thanks{e-mail: zhoulng@pku.edu.cn. This is the authors' version. L. Zhou is the corresponding author.}\\ %
     \parbox{2.2in}{   \scriptsize \centering Institute of Medical Technology, Peking University Health Science Center\\National Institute of Health Data Science, Peking University}} %

\abstract{%
 We present a new method to visualize data ensembles by constructing structured probabilistic representations in latent spaces, i.e., lower-dimensional representations of spatial data features.
  Our approach transforms the spatial features of an ensemble into a latent space through feature space conversion and unsupervised learning using a variational autoencoder (VAE). 
  The resulting latent spaces follow multivariate standard Gaussian distributions, enabling analytical computation of confidence intervals and density estimation of the probabilistic distribution that generates the data ensemble.
   Preliminary results on a weather forecasting ensemble demonstrate the effectiveness and versatility of our method.
}


\teaser{
  \centering
  \subfloat[Spaghetti plot]{\includegraphics[width=0.3\linewidth]{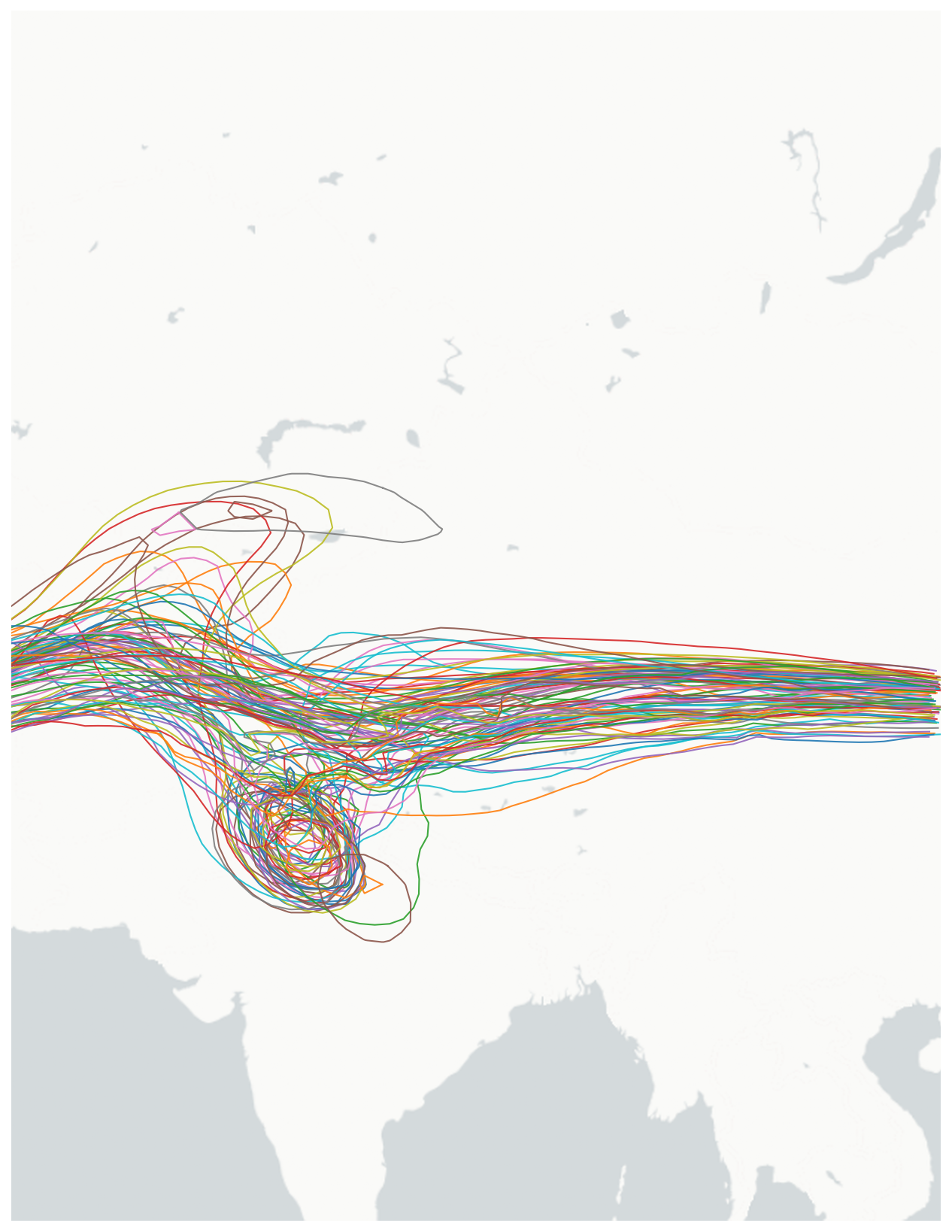}}\hfill
  \subfloat[Our method: probability density plot]{\includegraphics[width=0.3\linewidth]{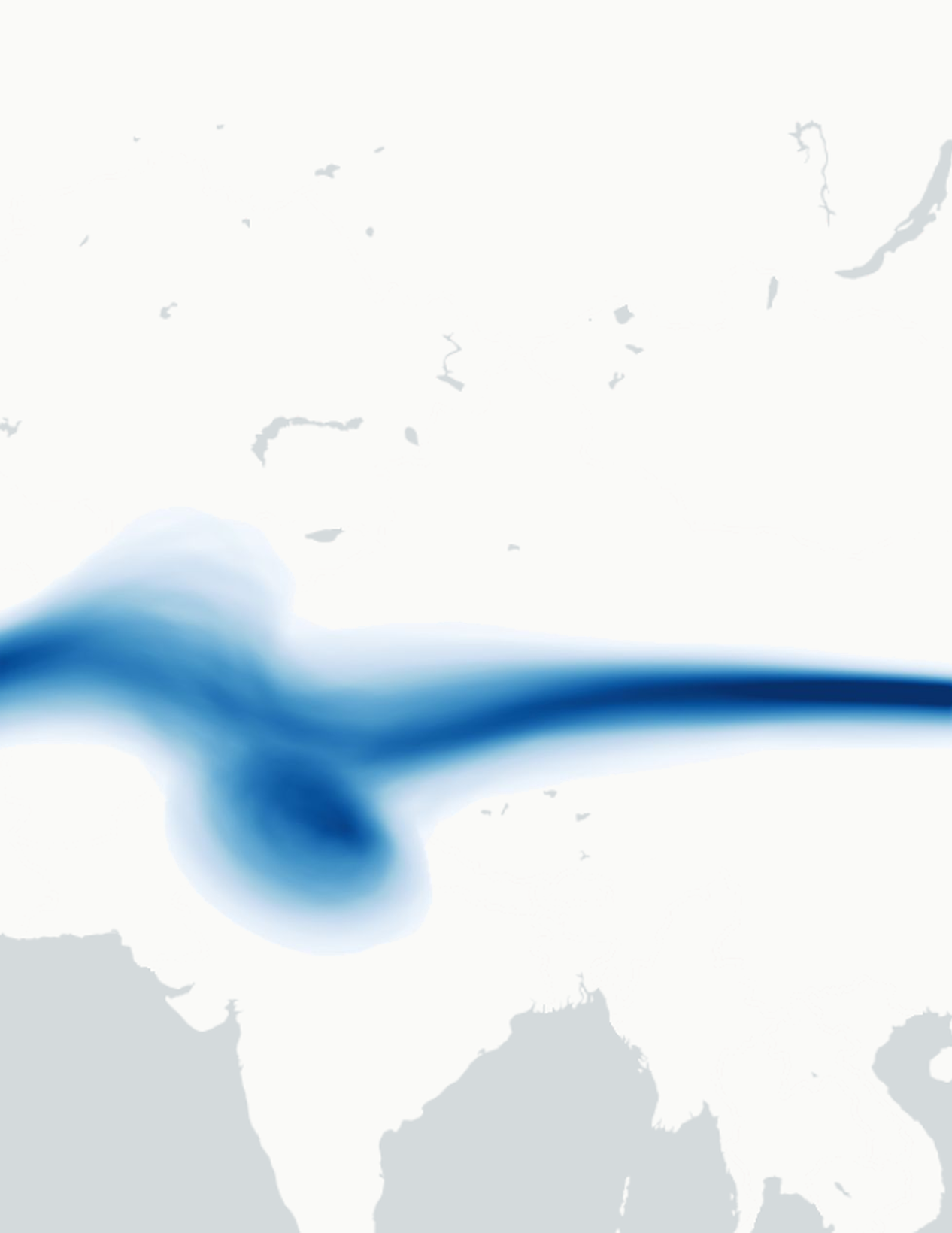}}\hfill
  \subfloat[Our method: confidence bands]{\includegraphics[width=0.3\linewidth]{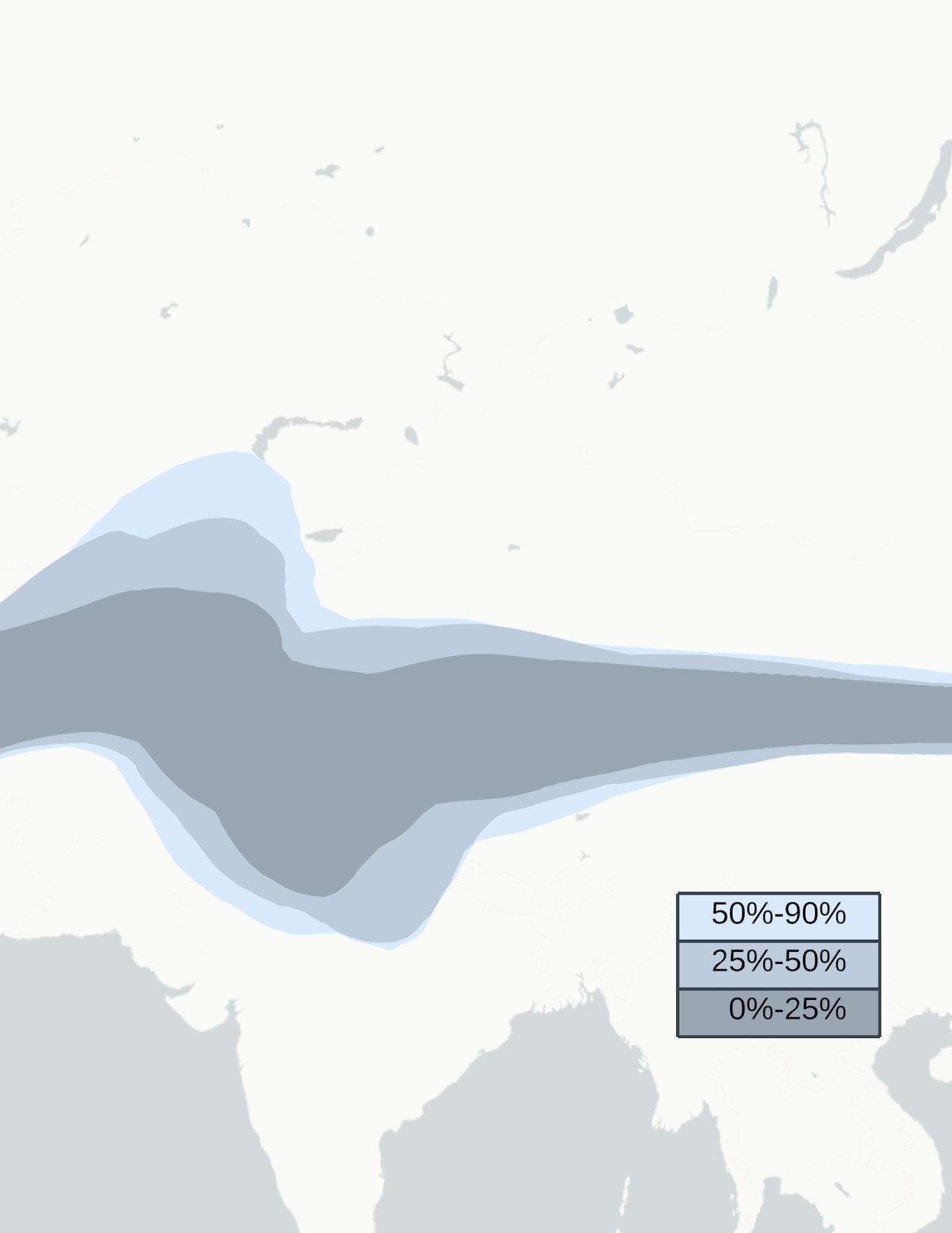}}\hfill
  \caption{%
 Visualizations of an ensemble from a weather simulation. 
 The spaghetti plot of ensemble members is shown in (a).
Ensemble visualizations of (b) the probability density plot and (c) confidence interval bands are supported by our method through computation in the VAE latent space.
  }
  \label{fig:teaser}
}

\graphicspath{{figs/}{figures/}{pictures/}{images/}{./}} 

\usepackage{tabu}                      
\usepackage{booktabs}                  
\usepackage{lipsum}                    
\usepackage{mwe}                       
\usepackage{amsfonts}
\usepackage{amsmath}
\usepackage{graphicx} 
\usepackage{float}
\usepackage{mathptmx}                  
\usepackage{subfig}
\usepackage{algorithm}
\usepackage{algpseudocode}

\begin{document}


\firstsection{Introduction}

\maketitle
Ensemble visualization is crucial for understanding uncertainty and variability in complex systems across domains such as weather forecasting, geophysics, and biomedical simulations. 
Spatial features such as contours and curves are fundamental in ensemble data analysis.

However, it is challenging to visualize these features effectively.
Spaghetti plots become cluttered with numerous ensemble members, obscuring important patterns.
Boxplot techniques using data depth represent statistics of ensemble members only~\cite{whitakerContourBoxplotsMethod2013,mirzargarCurveBoxplotGeneralization2014a}, but cannot show the probability density field.
Density-based methods try to visualize the probability distribution generating the spatial features~\cite{ferstlStreamlineVariabilityPlots2016,zhangEnConVisUnifiedFramework2023}.
However, they often rely on local statistical measures in the spatial domain or linear dimensionality reduction techniques having difficulty capturing complex  global structures. 
\begin{figure*}[tb] 
    \centering
    \includegraphics[width=\linewidth]{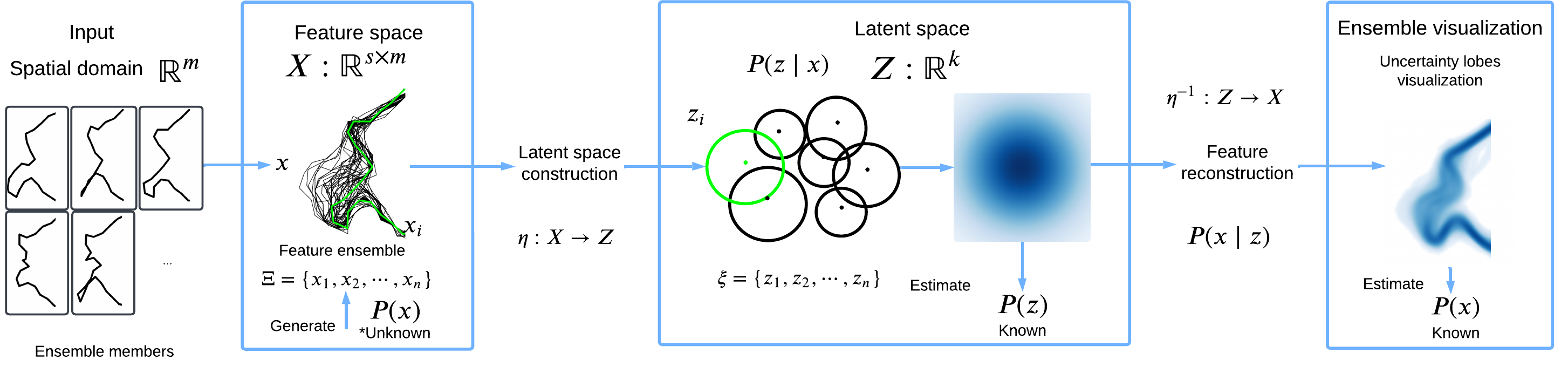}  
    \vspace{-1ex}
    \caption{The workflow of our method for ensemble visualization.} 
    \vspace{-1ex}
\label{fig:workflow}  
\end{figure*}

In this paper, we introduce a  method using the VAE~\cite{kingmaAutoencodingVariationalBayes2022} that creates structured, probabilistic representations of ensemble spatial features in latent spaces.
Our approach transforms spatial features into latent representations of lower dimensionalities that preserve both global structure and local variations while ensuring analytically computable probability distributions.
By exploiting the structured Gaussian properties of VAE latent spaces, our method supports multiple types of visualizations using a unified learned representation.
As shown in~\cref{fig:teaser}, our method can support ensemble visualization methods 
including probability density plots (\cref{fig:teaser}(b)), and confidence interval bands (\cref{fig:teaser}(c)).

The main contribution of our work is a VAE method for ensemble visualization that supports multiple types of visualizations through a single latent space representation. 
One benefit of our method is that the sampling of the latent space and the estimation of the probability distribution for ensemble members are facilitated with the properties of the Gaussian space learned by VAE.
Another benefit is that complex nonlinear spatial features are captured and can be reconstructed with high quality.

\section{Related Work}

Ensemble visualization is an active research topic~\cite{obermaierFutureChallengesEnsemble2014} and a core aspect of uncertainty visualization. 
Comprehensive surveys on ensemble~\cite{Wang2019,potterQuantificationVisualizationTaxonomy2012}. and uncertainty visualization can be found elsewhere~\cite{brodlieReviewUncertaintyData2012,dillExpandingFrontiersVisual2012,pangApproachesUncertaintyVisualization1997}.

Early works of ensemble visualization focus on direct statistical representation. 
Per-cell glyphs are employed to encode local statistics such as mean and variance~\cite{Kehrer2011,lodhaUFLOWVisualizingUncertainty,537309}, with various glyph designs developed for different data types~\cite{hlawatschFlowRadarGlyphs2011,sanyalNoodlesToolVisualization2010a}. 
While glyphs work well for sparsely sampled grids, they quickly lead to visual overload at higher resolutions. 
Spaghetti plots directly visualize structural features, such as contours, by overlaying all ensemble members over the context~\cite{Potter2009,sanyalNoodlesToolVisualization2010a,Zhang2021}.
However, visual clutter grows with the number of members with spaghetti plots.

Density-based approaches are available to address limitations of direct visualizations~\cite{grigoryanPointbasedProbabilisticSurfaces2004,pothkowProbabilisticMarchingCubes2011,poethkowApproximateLevelcrossingProbabilities2013a,scheepensInteractiveVisualizationMultivariate2011}.
Lampe and Hauser introduce curve density estimates, which transform large collections of 2-D function graphs into smooth probability maps that remain interpretable across zoom levels~\cite{lampeCurveDensityEstimates2011,Pfaffelmoser2013}. 
Extending this concept, He et al. introduce surface density estimates and the eFESTA framework, generalizing kernel density estimation from points to polygonal surface patches for ensembles of isosurfaces, ridgesurfaces, and streamsurfaces~\cite{heEFESTAEnsembleFeature2019}. 
Kumpf et al. compute point-wise contour probabilities on signed-distance grids~\cite{kumpfVisualizingConfidenceClusterbased2018,demirVisualizingCentralTendency2016}. 
The signed distance is further extended using kernel density estimation as a unified approach for several existing density-based methods~\cite{zhangEnConVisUnifiedFramework2023}.  
These density-field approaches maintain spatial coherence while reducing the clutter inherent in dense data.
Unlike these methods that rely on local probability estimation, our approach approximates the overall probability density distribution with the nonlinear latent space learned by VAE.

Dimensionality-reduction techniques provide another avenue for ensemble analysis by embedding features in low-dimensional sub-spaces. 
Ferstl et al. derive variability plots for streamlines and contours via PCA spaces~\cite{ferstlStreamlineVariabilityPlots2016,ferstlVisualAnalysisSpatial2016}.  
Popular nonlinear dimensionality reduction methods, for example, t-distributed stochastic neighbor embedding (t-SNE)~\cite{vanderMaaten08a}, and uniform manifold approximation and projection (UMAP)~\cite{McInnes2018} generate nonlinear latent spaces that can describe complex features.
However, these latent spaces are typically not invertible, i.e., cannot be used for ensemble feature reconstruction, and they are not regular and cannot be used for analytical computations.
Our VAE method is also a form of nonlinear dimensionality reduction but addresses these limitations to support ensemble visualization.

\section{Methods}
\label{sec:methods}
In this section, we present our computational method for ensemble visualization and explain its key components in detail. 
Our method consists of three stages: feature space conversion, latent space construction, and uncertainty visualization.

The dataset used in this paper is a weather simulation generated by the ensemble prediction system of the European Centre for Medium-Range Weather Forecasts (ECMWF)\footnote{\url{https://www.ecmwf.int}}.
We adopt an ensemble of 95 members of a geopotential height field at 500 hPa within the East Asia region~\cite{zhangEnConVisUnifiedFramework2023}.

\subsection{Probabilistic Latent Space Modeling with VAE}
\label{sec:vae_method}
Our method employs a VAE to construct a probabilistic latent space to model the ensemble data. 
The process involves two steps: converting spatial features to a high-dimensional feature space (\cref{fig:workflow}-Feature space), and learning a structured latent space from the feature space (\cref{fig:workflow}-Latent space).

\subsubsection{Feature Space Conversion}
We convert ensemble members from the spatial domain to a feature space through arc-length parameterization.
For each of the $n$ contours in the $m$-dimensional spatial domain, we uniformly sample $s$ points along its arc-length and record their $m$-dimensional coordinates.
This transforms each contour into a feature vector of size $sm$.
The resulting ensemble is $\Xi=\{x_1,x_2,\cdots,x_n \}$, where each $x_i \in \mathbb{R}^{s \times m}$ represents a feature vector in the feature space $X = \mathbb{R}^{s \times m}$.
This approach ensures equal representation of all contour parts while providing a compact representation that captures the geometric characteristics of features and leads to stable training for the VAE.

\subsubsection{The Training of VAE}
The VAE consists of an encoder that maps $x$ in feature space $X$ to $z$ in a latent space $Z$ of $\mathbb{R}^k$ ($k < sm$) and a decoder that reconstructs features back to $X$ from $Z$~\cite{kingmaAutoencodingVariationalBayes2022}. 
We set $k=8$ as the latent dimension, as our experiments on the weather data show that this setting provides a good balance between the compactness of dimensionality and the reconstruction quality.
The mapping follows the Bayesian theorem:
\begin{equation}
    p(z|x) = \frac{p(x|z)p(z)}{p(x)}\;,
    \label{eqn:basemodel}
\end{equation}
where $p(z)$ is the prior (standard Gaussian), $p(z|x)$ is the posterior approximated by $q(z|x)$, and $p(x|z)$ is the reconstruction likelihood. 
The VAE is trained using the objective function:
\begin{equation}
    \mathcal{L} = \mathbb{E}_{q(z\mid x)}[\log p(x\mid z)] - D_{KL}(q(z\mid x)||p(z))\;,
\end{equation}
where $D_{KL}$ is the Kullback-Leibler divergence. 
This loss function balances reconstruction fidelity (the first term) with regularization that enforces the latent space to follow a standard Gaussian distribution (the second term).
The formulation of VAE results in several properties facilitating the probabilistic modeling of an ensemble.

\begin{figure*}[htb] 
    \centering
    \subfloat[Ensemble members]{\includegraphics[trim={2cm 0 2cm 0}, clip, width=0.24\linewidth]{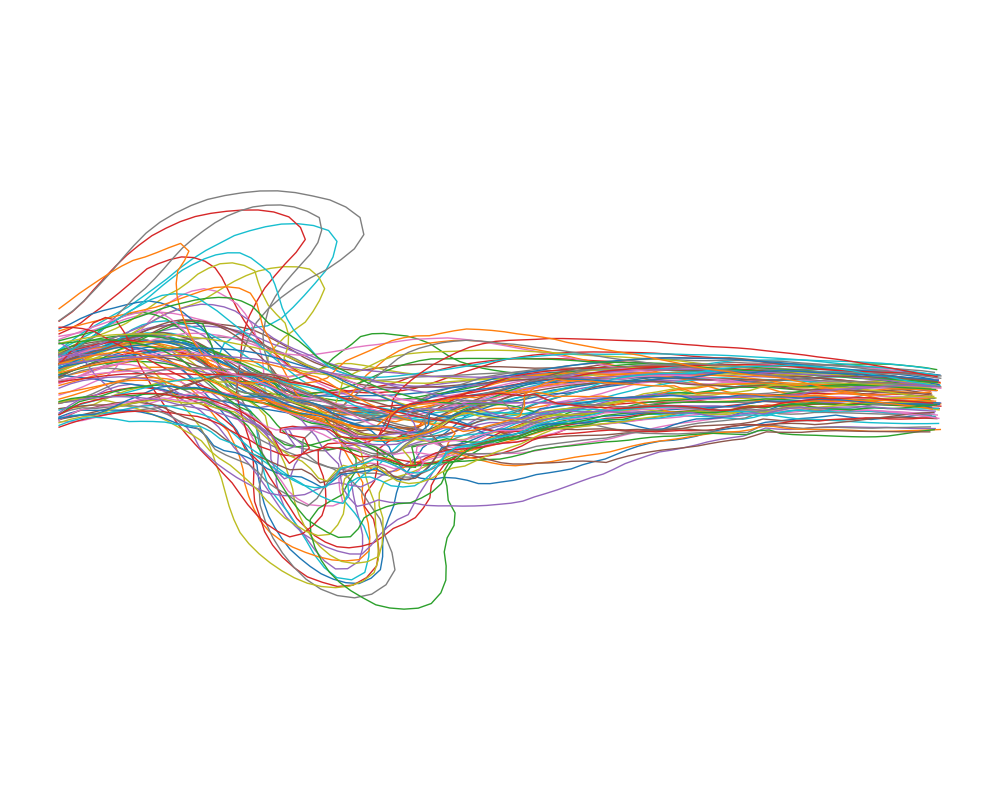}}\hfill
    \subfloat[Reconstructed (neighborhood sampling) ]{\includegraphics[trim={2cm 0 2cm 0}, clip, width=0.24\linewidth]{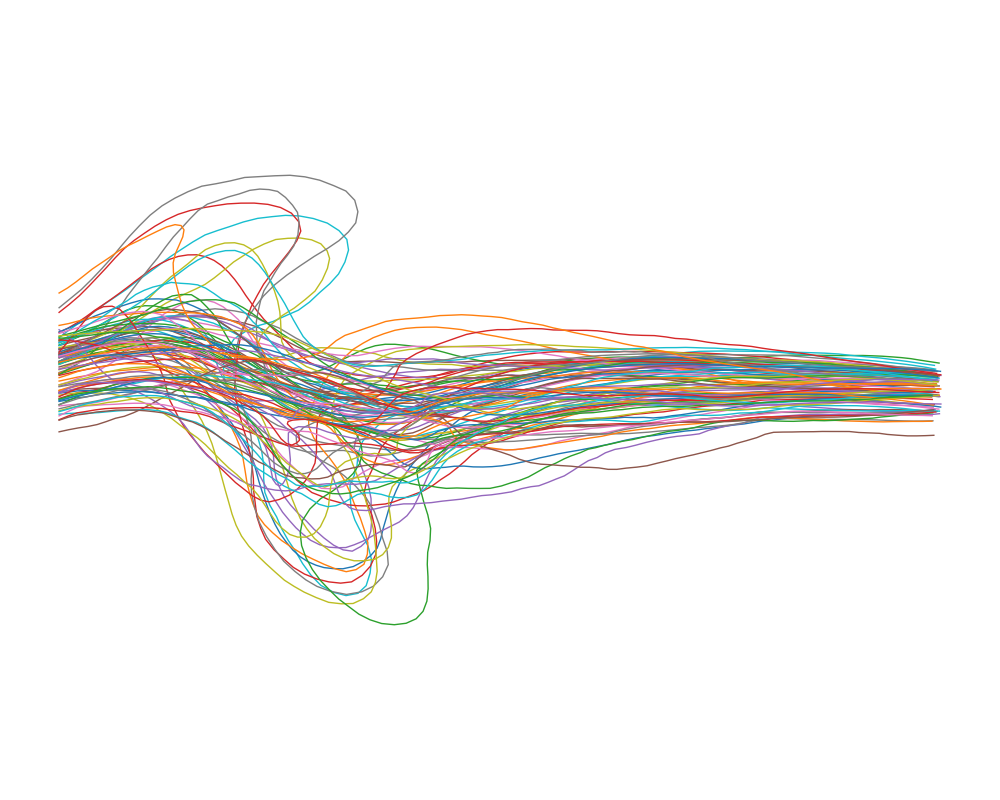}} \hfill
    \subfloat[Reconstructed (importance sampling)]{\includegraphics[trim={2cm 0 2cm 0}, clip, width=0.24\linewidth]{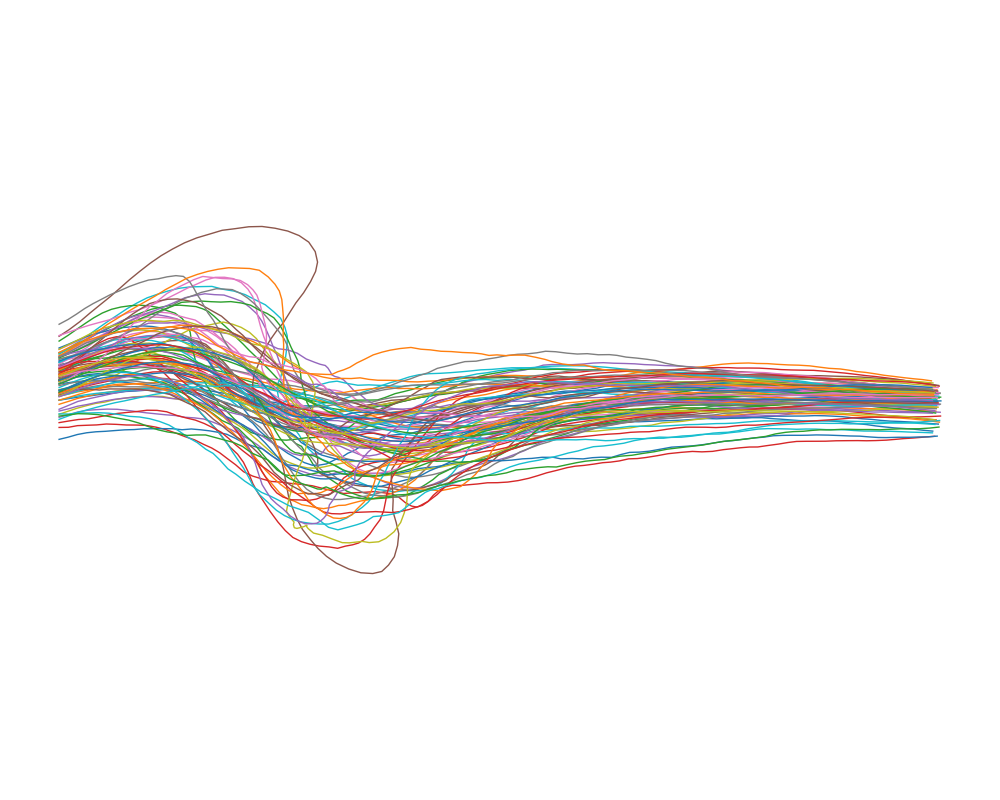}}  \hfill
    \subfloat[Reconstructed (importance sampling with 300 samples)]{\includegraphics[trim={2cm 0 2cm 0}, clip, width=0.24\linewidth]{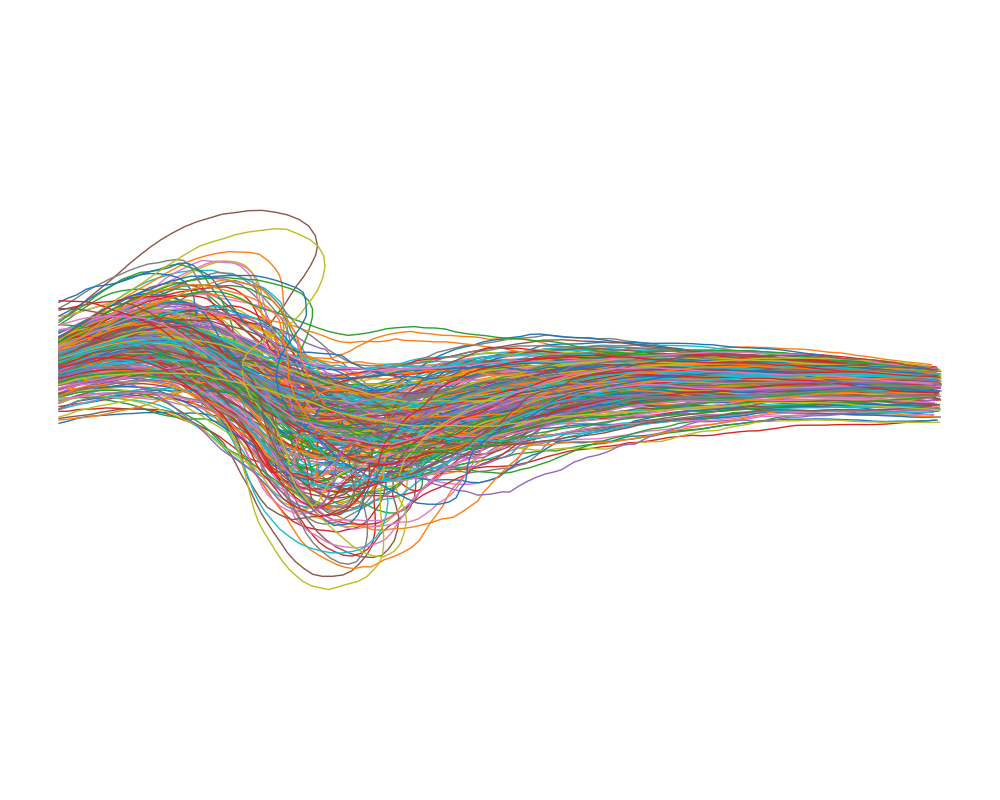}}  
    \caption{A comparison of (a) 95 original data samples and contours generated from the latent space of VAE (b--d): (b) 95 samples in the neighborhood of original data, (c) 95 samples according to Gaussian distribution, and (d) 300 samples according to Gaussian distribution.}
    \label{fig:generatedSamples}  
\end{figure*}

\subsubsection{Properties of the Latent Space}
The trained VAE produces a structured latent space where data follow a multivariate standard Gaussian distribution with properties that are important for effective uncertainty visualization.

\textbf{Reconstruction Fidelity.}
Important spatial characteristics are largely preserved during the encoding-decoding process of VAE~\cite{kingmaAutoencodingVariationalBayes2022}. 
This is crucial for our approach as it helps ensure that uncertainty visualizations generated from the latent space reasonably reflect the variability of the original ensemble data. \cref{fig:generatedSamples} shows the fidelity of the reconstruction by comparing the original data samples (\cref{fig:generatedSamples}(a)) with samples generated from the latent space (\cref{fig:generatedSamples}(b-d)).

\textbf{Continuity and Interpolation.}
Building upon the reconstruction fidelity, the VAE latent space exhibits continuity, where nearby points in $Z$ correspond to similar features in the original space $X$~\cite{mi2021revisiting}. 
This enables meaningful interpolation between ensemble members, allowing for the generation of intermediate features that maintain physical plausibility and ensure smooth transitions in uncertainty visualizations. 
As demonstrated in \cref{fig:generatedSamples}(b), contours generated by sampling the neighborhood of ensemble members are similar to the original members (\cref{fig:generatedSamples}(a)).
\cref{fig:generatedSamples}(c, d) show how a wider coverage of the sampling according to the Gaussian distribution maintains overall trends while introducing reasonable variations.

\textbf{Analytical Computation.}
The multivariate standard Gaussian distribution of the latent space enables analytical computation of confidence intervals and probability measures. 
The latent space allows for efficient importance sampling and precise determination of confidence regions using properties of the chi-square distribution for uncertainty quantification.

\subsection{Ensemble Visualization}
\label{sec:uncertainLobes}
\begin{figure}[htb] 
    \centering
    \includegraphics[width=\linewidth]{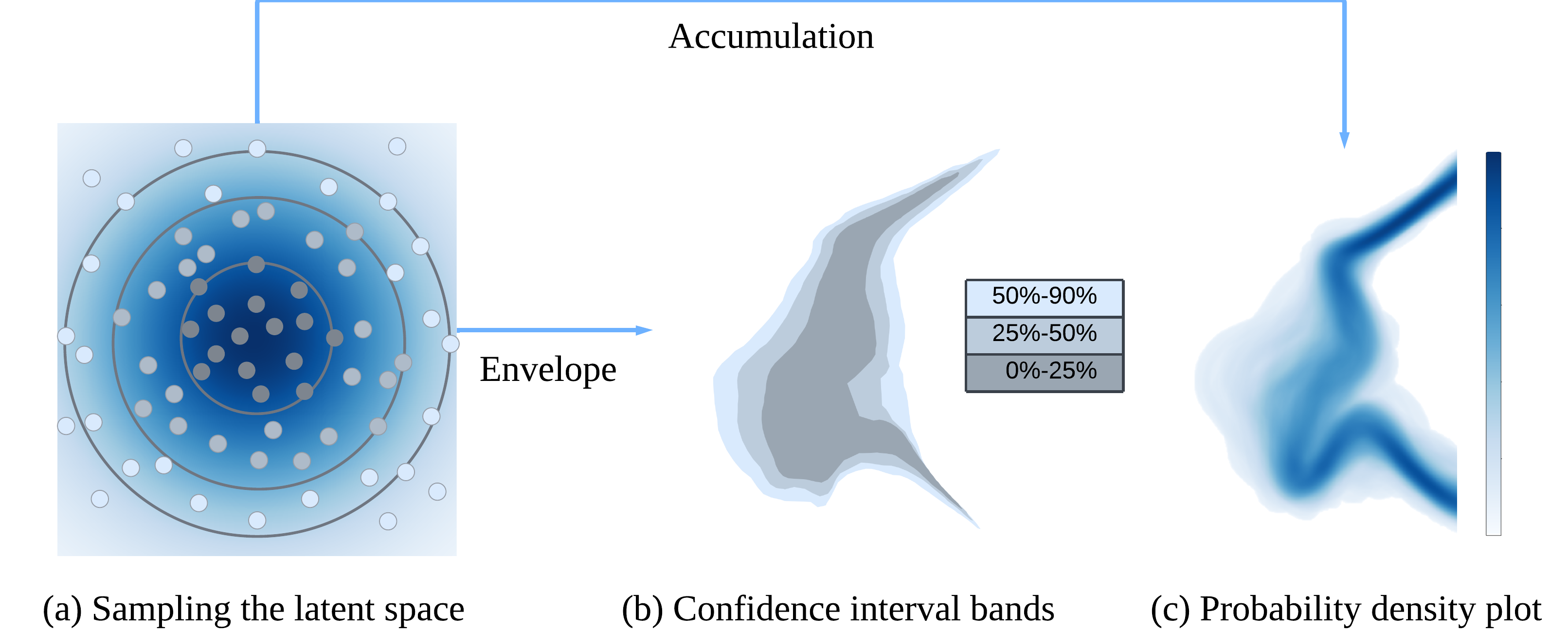}    
    \caption{Uncertainty lobes are characterized by (a) sampling the latent space--illustrated as a 2D Gaussian distribution, and feature reconstruction. 
    With samples placed within radii of spheres (dots of different shades of blue-gray) associated with given confidence levels, the confidence interval bands is visualized as the envelope of the associated features of the samples (b). 
    By sampling the whole latent space (all dots), (c) the probability density function of the ensemble is visualized as a probability density plot through accumulation. }
    \label{fig:uncertainLobes}  
\end{figure}
Uncertainty lobes~\cite{ferstlStreamlineVariabilityPlots2016,ferstlVisualAnalysisSpatial2016} and probability density plots~\cite{kumpfVisualizingConfidenceClusterbased2018,zhangEnConVisUnifiedFramework2023} are effective means for visualizing ensemble datasets.
In our method, we approximate the probability distribution $P(x)$ by the probability $p(x)$.
Instead of directly estimating $p(x)$, we use the latent space to approximate $p(x)$ with $p(z)$ through $p(x\mid z)$ in our approach (\cref{fig:workflow}-Feature reconstruction).

Specifically, our VAE approach estimates the probability density function and confidence intervals in the latent space \(Z\) and then uses the decoder to generate the corresponding uncertainty signatures in the original feature space \(X\) as shown in~\cref{fig:uncertainLobes}.
Because of the aforementioned properties of VAE, the structures reconstructed from \(Z\) can serve as a good approximation of that in \(X\). 
Furthermore, 
the confidence intervals can be determined analytically, improving estimation accuracy.
\begin{figure*}[tb]
    \centering
    \subfloat[Probability density plot based on PCA-CVP~\cite{ferstlStreamlineVariabilityPlots2016}]{%
        \includegraphics[width=0.33\textwidth]{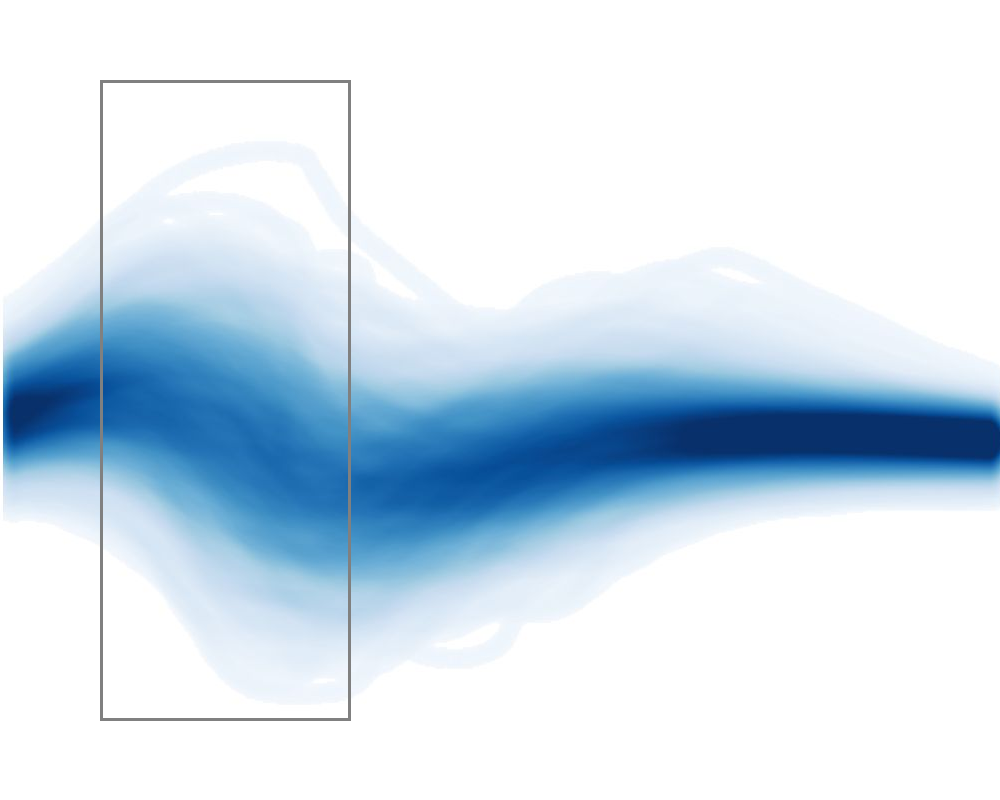}
    }
    \subfloat[Spaghetti plot of member contours]{%
        \includegraphics[width=0.33\textwidth]{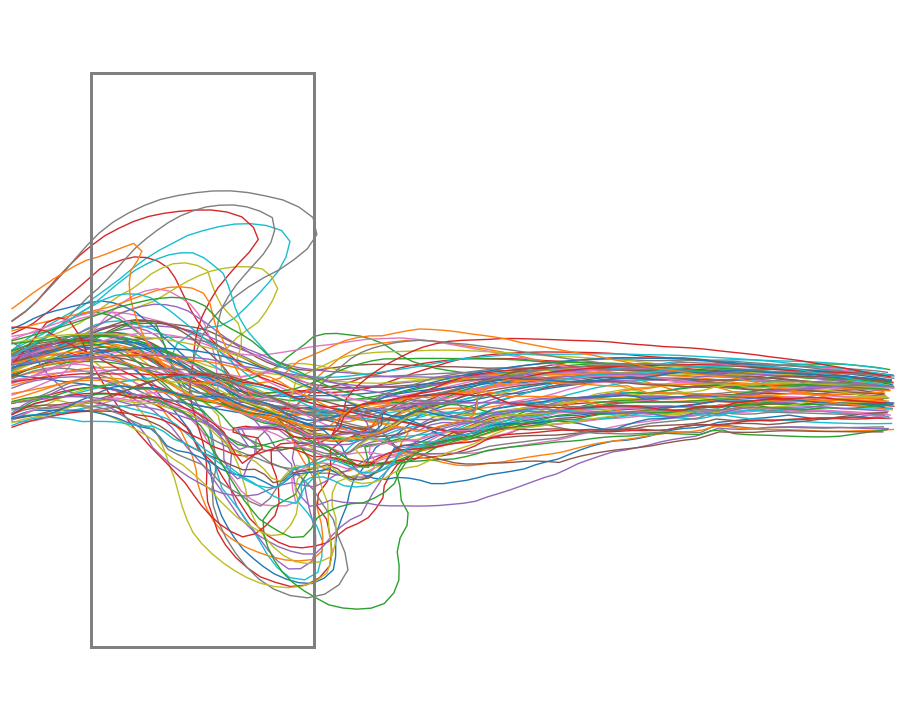}
    }
    \subfloat[Probability density plot with our method]{%
        \includegraphics[width=0.33\textwidth]{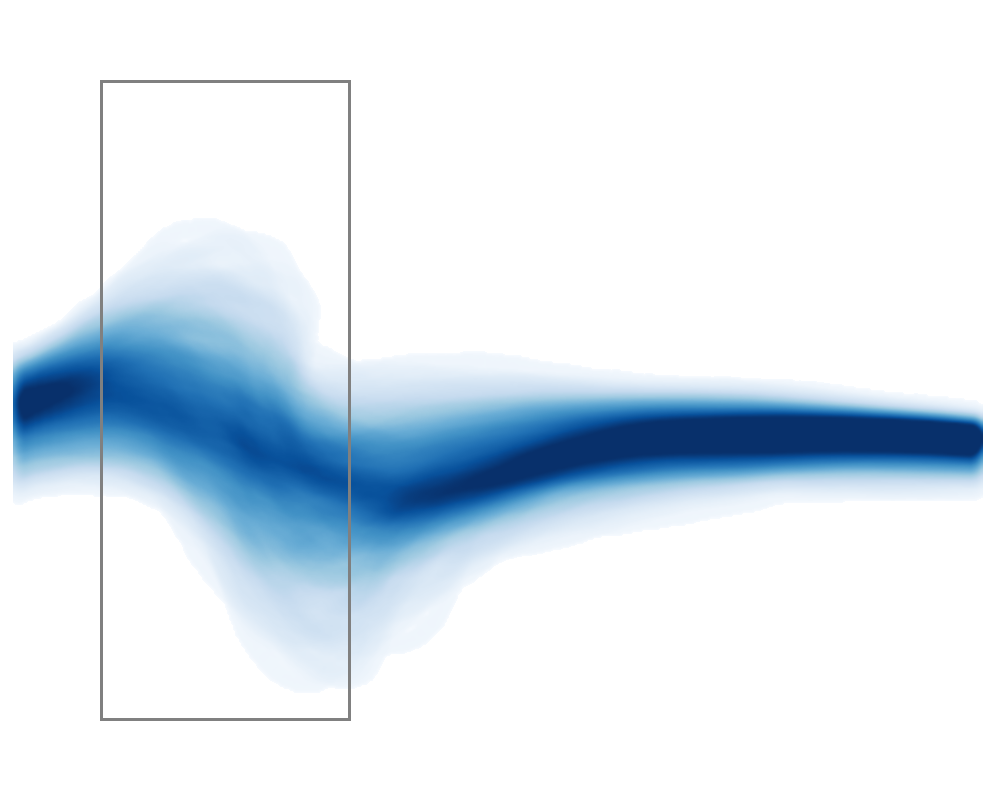}
    }
    \caption{A comparison of density estimation approaches for (b) the ensemble members shown in a spaghetti plot. (a) Probability density plot based on the PCA-CVP method; (c) Probability density plot of our method.}
    \label{fig:SupplDensityComparison}
\end{figure*}
\subsubsection{Confidence Interval Bands}

The latent space \(Z\) is a multivariate Gaussian distribution whose probability density function reads: 
\begin{equation}
  p(z) = \frac{1}{(2\pi)^{k/2}} \exp\left(-\frac{1}{2}\|z\|^2\right)\;.
\end{equation}
For the latent space $Z$, the squared Euclidean distance from the origin \(\|z\|^2 = z_1^2 + z_2^2 + \cdots + z_k^2\) follows a chi-square distribution \(\chi^2_k\) with \(k\) degrees of freedom. 
This is because that each component \(z_i\) is independently and identically distributed as a standard normal random variable, and the sum of squares of \(k\) independent standard normal variables follows a \(\chi^2_k\) distribution.

Therefore, a confidence interval region in \(Z\) is a $k$-dimensional sphere centered at the mean (origin) of the Gaussian.
For a given confidence level \(1-\alpha\), we need to find the radius \(r\) such that:
\begin{equation}
  P(\|z\|^2 \leq r^2) = 1-\alpha\;,
\end{equation}
the corresponding radius \(r\) is determined by:
\begin{equation}
  r = \sqrt{\chi^2_{k,1-\alpha}}\;,
\end{equation}
where \(\chi^2_{k, 1-\alpha}\) is the chi-square critical value with \(k\) degrees of freedom.
We then uniformly sample the confidence sphere of radius $r$ in the latent space $Z$ (one shade of blue-gray dots of \cref{fig:uncertainLobes}(a)), and reconstruct the samples in the original feature space $X$  through the decoder of the VAE. 
By taking the envelope of all these contours, we obtain the confidence interval band in \(X\) corresponding to the chosen confidence level $1-\alpha$ (\cref{fig:uncertainLobes}(b)).

Unlike point-based techniques~\cite{kumpfVisualizingConfidenceClusterbased2018, zhangEnConVisUnifiedFramework2023}, our method is feature-based and a quantified version of the variability plots~\cite{ferstlStreamlineVariabilityPlots2016} thanks to the latent space of VAE.
Therefore, the confidence interval bands illustrate the global instead of local properties of ensemble members in the spatial domain.
\cref{fig:teaser}(c) and \cref{fig:uncertainLobes}(b) show visualizations of composited confidence bands calculated for \(1-\alpha\) = 0.25, 0.5, and 0.9.
In this paper, we use $1000$ sample points to cover the sphere as it shows plausible variations.

\subsubsection{Probability Density Plots}

It is also of interest to obtain a detailed representation of the probability distribution of contours in the original feature space \(X\).
To achieve this, we use importance sampling to sample the entire latent space \(Z\) according to the standard Gaussian distribution (all dots in \cref{fig:uncertainLobes}(a)).
With a sufficient number of samples, the accumulated frequency of these decoded contours approximates the true distribution of contours in the feature space \(X\), and, subsequently, the spatial domain.
The density plots visualize the uncertainty inherent in the ensemble with high-density regions indicating a higher likelihood of the occurrence of a feature, e.g., how likely a feature may pass through.
Example density plots of the approximated density distribution are shown in ~\cref{fig:teaser}(b) and~\cref{fig:uncertainLobes}(c) with white (minimum) to blue (maximum) ColorBrewer colormaps.

\section{Results}
Visualization results of our method with the weather ensemble overlaid on the regional map are shown in~\cref{fig:teaser}.
The probability density plot shows the fine structures of the probability distribution of the height field in the spatial domain (\cref{fig:teaser}(b)).
The confidence bands convey the coverage of contours at different confidence intervals (\cref{fig:teaser}(c)).

A side-by-side comparison of three visualizations of the main connected contours of the weather data is shown in~\cref{fig:SupplDensityComparison}: the probability density plot by PCA-contour variability plot (PCA-CVP)~\cite{ferstlStreamlineVariabilityPlots2016}  (\cref{fig:SupplDensityComparison}(a)), the spaghetti plot of the original data (\cref{fig:SupplDensityComparison}(b)), and the probability density plot with our method (\cref{fig:SupplDensityComparison}(c)).
The comparison, especially for the part highlighted in the gray box,  reveals that the PCA-CVP  (\cref{fig:SupplDensityComparison}(a)) captures contours with relatively small variations but has difficulty capturing the density of contours having big variations as those crossing the main trends top-down that can be seen in the spaghetti plot (\cref{fig:SupplDensityComparison}(b)). 
This limitation stems from the constraints of PCA in capturing complex nonlinear variations in the data.

In contrast, our VAE-based reconstruction (\cref{fig:SupplDensityComparison}(c)) correctly captures these high-variation regions and provides a better representation of the original data variability shown in the spaghetti plot compared to the PCA method.

\begin{table}[h]
\centering
\begin{tabu}{lc}
\toprule
Method  & MMD-CD \\
\midrule
PCA (CVP)  & 0.663 \\
VAE (Ours)  &\textbf{0.578} \\
\bottomrule
\end{tabu}
\caption{Minimum Matching Distance-Chamfer Distance (MMD-CD) between 95 generated and original contours of the weather data. Lower values indicate higher similarities to the original data.}
\label{tab:mmdcd_comparison}
\end{table}

A preliminary quantitative evaluation further supports this observation. 
We generate 95 contours sampled according to the Gaussian distribution of the VAE latent space (8-D standard Gaussian), and 95 contours according to a Gaussian fitted to the 8-D PCA embedding~\cite{ferstlStreamlineVariabilityPlots2016}, respectively.
The two generated datasets are compared against the original 95 members using Minimum Matching Distance-Chamfer Distance (MMD-CD)~\cite{pmlr-v80-achlioptas18a}. 
By uniformly sampling points on each contour and computing the bidirectional Chamfer distance between point sets, and, for each reference contour, the best matching is the minimum distance to any generated contour. 
MMD-CD is then the average of these minima across references, and a lower score corresponds to higher similarity. 
As shown in~\cref{tab:mmdcd_comparison}, VAE achieves a $12.77\%$ reduction in MMD-CD compared to PCA. 
This suggests that compared to PCA, the VAE captures the global structure of the ensemble better and generates contours with higher similarity to original members.

\section{Conclusion}
We have presented a VAE method for ensemble visualization that addresses some limitations of existing approaches through structured probabilistic representations in latent spaces.
By leveraging the structured Gaussian properties of VAE latent spaces, our approach enables the generation of probability density plots and confidence interval bands from a unified learned representation.
Unlike traditional methods that rely on local probability estimation or linear dimensionality reduction, our approach approximates overall probability density distributions through sampling and reconstruction in the latent space.
Preliminary results indicate improvements over PCA-based methods in capturing complex nonlinear variations present in spatial ensemble features and providing enhanced uncertainty quantification through analytical computation of confidence intervals.

For future work, we would like to extend for 3D curve and contour ensembles.
Another direction is to support visual analysis to better understand the latent space.
A comprehensive quantitative evaluation of various latent spaces for ensemble visualization is also of our interest.

\acknowledgments{%
  This work was supported by the NSFC grant (No.~62372012).%
}

\bibliographystyle{abbrv-doi-hyperref}

\bibliography{datadepth}

\appendix

\end{document}